\pdfoutput=1
\documentclass[letterpaper]{article}
\usepackage[preprint]{aaai2027}
\usepackage[hyphens]{url}
\usepackage{graphicx}
\usepackage{natbib}
\usepackage{caption}
\usepackage{subcaption}
\usepackage{algorithm}
\usepackage{algorithmic}

\usepackage{newfloat}
\usepackage{amsthm}

\usepackage{amsfonts}
\usepackage{amsmath}
\usepackage{listings}
\usepackage{soul}
\DeclareCaptionStyle{ruled}{labelfont=normalfont,labelsep=colon,strut=off}
\floatstyle{ruled}
\newfloat{listing}{tb}{lst}{}
\floatname{listing}{Listing}

\usepackage{booktabs}

\title{Invertible Transport Between Sample-Only Distributions\\ via Projected Characteristic Functions}
\author{
    Juan Gutiérrez\textsuperscript{\rm 1,\rm 2},
    Victor Gutiérrez-García\textsuperscript{\rm 2},
    José Luis Blanco-Murillo\textsuperscript{\rm 1,\rm 2}
}
\affiliations{
    \textsuperscript{\rm 1}Information Processing and Telecommunications Center\\
    \textsuperscript{\rm 2}ETSI de Telecomunicación, Universidad Politécnica de Madrid, Spain\\
    \{juan.gutierrez, v.ggarcia, jl.blanco\}@upm.es
}

\begin{document}

\maketitle
\thispagestyle{plain}

\begin{abstract}
Many useful latent distributions can be sampled but do not have a tractable density. Such distributions may be represented by empirical collections of real-world observations or generated by scientific simulators and black-box generative models. These sample-only priors cannot be used directly in likelihood-based normalizing flows, which require evaluating the prior density. We propose Projected Characteristic-Function matching (PCF), a density-free framework for learning an invertible map between a data distribution and a prior specified only through samples.

Our method aligns distributions using projected empirical characteristic functions, eliminating the need for density estimation, adversarial training, or paired observations. To address the non-uniqueness of the transport map, we introduce an optional transport regularizer that selects lower-displacement transformations while preserving endpoint alignment.

We benchmark PCF on controlled transports, sample-efficiency studies, and high-dimensional alignment tasks. PCF matches the accuracy of kernel MMD on the controlled benchmark and on raw-pixel MNIST--USPS at a cost linear in the batch size, and it is more sample-efficient, achieving lower endpoint error than MMD in 13 of 15 paired runs across library sizes. By avoiding pairwise kernel evaluation, it reduces discrepancy memory by 29.8\% and trains at batch sizes where kernel MMD exceeds GPU memory. These results show that PCF provides accurate and efficient invertible transport, particularly in sample- and memory-limited settings.
\end{abstract}

\section{Introduction}
\label{sec:introduction}

Learning a transformation between two probability distributions from finite, unpaired samples is a fundamental problem in generative modeling, domain translation, simulation-based inference, and representation learning. In many of these settings, it is desirable for the transformation to be invertible: samples should be transported in both directions, and every generated observation should admit an exact representation in the source space. Normalizing flows provide a natural framework for learning such transformations through compositions of invertible functions or continuous-time dynamics \citep{dinh2016realnvp,kingma2018glow,chen2018neuralode,grathwohl2019ffjord,papamakarios2021normalizingflows}.

The standard normalizing-flow formulation, however, is designed around a tractable base distribution, most commonly an isotropic Gaussian. Maximum-likelihood training evaluates the density of an encoded observation under this base law together with the change-of-variables correction. This construction is effective when the base distribution is chosen for computational convenience, but it does not directly address settings in which the latent distribution is externally prescribed and accessible only through samples. Such sample-only priors can arise from simulators, collections of trajectories, empirical latent libraries, molecular samplers, physical state distributions, or black-box generative models. In these cases, the structure of the prior carries information: it defines the coordinate law in which observations should be represented, rather than merely supplying unstructured noise for generation.

Let $P_{\mathcal X}$ denote the data distribution and let $P_{\mathcal Z}$ denote a prescribed prior distribution. We assume access to independent samples from both distributions, but do not assume that the density of $P_{\mathcal Z}$ can be evaluated. Our goal is to learn an invertible map $F_\theta$ with inverse $G_\theta=F_\theta^{-1}$ such that
\begin{equation}
(F_\theta)_\# P_{\mathcal X}=P_{\mathcal Z},
\qquad
(G_\theta)_\# P_{\mathcal Z}=P_{\mathcal X}.
\label{eq:intro-endpoints}
\end{equation}
Thus, encoded observations should follow the prescribed latent law, while prior samples should decode into the data distribution. Although the two equalities are equivalent for an exact invertible map at the population level, enforcing both during stochastic optimization provides direct training signals in both directions.

Recent work has expanded invertible generative modeling beyond conventional maximum-likelihood training. Sample-based objectives, including sliced-Wasserstein discrepancies, energy and proper-scoring-rule losses, and score matching, show that normalizing flows can be trained without relying exclusively on tractable likelihoods \citep{coeurdoux2022sliced,chao2023ebflow}. However, these approaches typically retain an analytically specified base distribution or focus on modeling a single target law. Fitting an auxiliary density model to samples from the desired prior offers one alternative \citep{golling2023morphing}, but introduces an additional estimation stage whose errors can affect the learned transport. Path-based approaches such as Flow Matching instead learn dynamics along a chosen probability path between the endpoints \citep{lipman2023flowmatching}, while drifting methods construct distributional dynamics from discrepancies or gradient flows \citep{gretton2026drifting}. Although these methods can operate with samples from both distributions, they require an interpolation, coupling, or drift construction beyond the endpoint laws themselves. Flow Matching, in particular, learns the endpoint laws only indirectly through coupled samples; with the same network, solver, and training budget, its endpoint error is 2.5--3.7 times that of PCF in our controlled benchmark.

Our starting point is LeJEPA, which introduces Sketched Isotropic Gaussian Regularization to match learned representations to an isotropic Gaussian through projected empirical characteristic functions \citep{balestriero2025lejepa}. This construction combines random one-dimensional projections with characteristic-function comparisons, providing a scalable sample-based mechanism for controlling a high-dimensional distribution. We extend this mechanism from representation regularization toward a fixed Gaussian law to invertible generative modeling between general sample-defined endpoint distributions. The proposed objective matches encoded data to independent prior samples and decoded prior samples to independent data samples, requiring neither prior-density evaluation, adversarial training, nor source--target pairings. We study both an unregularized formulation, which enforces the endpoint distributions, and a transport-regularized formulation, which additionally favors lower-cost invertible maps.

The approach is motivated by the fact that characteristic functions uniquely characterize probability distributions and by the Cramér--Wold principle, which reduces multivariate distribution matching to comparisons of one-dimensional projections \citep{cramer1936some,epps1983test}. Random directions and a multiscale frequency grid yield a differentiable minibatch objective based on bounded Fourier features \citep{ansari2020characteristic,knop2020cramerwold}. We state endpoint identifiability for the idealized all-direction, all-frequency discrepancy and evaluate its practical finite-grid surrogate against alternative endpoint, path-based, and likelihood-based objectives.

Our main contributions are:
\begin{itemize}
    \item We formulate prescribed sample-only prior learning as
    bidirectional endpoint matching with a single invertible map.
    \item We introduce PCF, a density-free two-sample objective that extends
    projected characteristic-function regularization to arbitrary endpoint
    distributions, with optional transport-cost regularization.
    \item We establish endpoint identifiability for the idealized discrepancy
    and compare its finite-grid surrogate with endpoint-, path-, and
    likelihood-based alternatives.
\end{itemize}

\section{Related Work}
\label{sec:related-work}

\paragraph{Likelihood-based and likelihood-free invertible models.}
Normalizing flows learn tractable invertible transformations between a
reference law and a data distribution, with maximum-likelihood training
enabled by the change-of-variables formula
\citep{dinh2016realnvp,kingma2018glow,chen2018neuralode,grathwohl2019ffjord}.
Their invertibility has also been exploited for unpaired domain
translation through shared latent representations and cycle consistency
induced by invertibility \citep{grover2020alignflow}. Subsequent work has reduced
dependence on likelihood through sliced-Wasserstein matching
\citep{coeurdoux2022sliced}, energy and proper-scoring-rule objectives and score matching \citep{chao2023ebflow}.
Another strategy fits an auxiliary density model before learning the
transformation \citep{golling2023morphing}. These
approaches broaden the training objectives for invertible
models, but generally retain an analytically specified reference
distribution or rely on an auxiliary density model.

\paragraph{Path-based and sample-to-sample transport.}
Continuous-time generative modeling has also developed toward
simulation-free path learning. Flow Matching trains vector fields along
prescribed conditional probability paths \citep{lipman2023flowmatching},
while minibatch and optimal-transport couplings can produce straighter
or lower-cost trajectories \citep{pooladian2023multisample}. Symmetric
MMD objectives combine bidirectional endpoint matching with transport
regularization to learn invertible maps directly from samples
\citep{xiong2023symotflow}. Related developments extend flow-based
training to constrained and function-valued domains
\citep{xie2024reflected,kerrigan2024functional}, conditional
distribution transfer \citep{wahl2025trade}, and flow-map distillation
\citep{sabour2025alignyourflow}. Drifting methods provide another
sample-based perspective by evolving particles through distributional
velocity fields \citep{deng2026drifting}, with recent analyses
connecting these dynamics to Wasserstein gradient flows associated with
kernel, Sinkhorn, and sliced-Wasserstein discrepancies
\citep{gretton2026drifting}. In contrast, our objective directly matches
the two endpoint laws without introducing a probability path or
coupling.

\paragraph{Characteristic-function distribution matching.}
A complementary line of work characterizes distributions through
one-dimensional projections and characteristic functions. The
Cramér--Wold principle and empirical characteristic-function tests
provide the classical foundations
\citep{cramer1936some,epps1983test}. These ideas have been used for
latent distribution matching in autoencoders
\citep{knop2020cramerwold}, implicit generative modeling through
weighted characteristic-function distances
\citep{ansari2020characteristic}, and joint distribution learning
\citep{an2023jointcw,an2023mixturecw}. More recently, LeJEPA introduced
a scalable projected characteristic-function regularizer for matching
high-dimensional representations to an isotropic Gaussian
\citep{balestriero2025lejepa}, with subsequent work relating this
construction to sliced kernel discrepancies and simplified statistical
variants \citep{zimmermann2025kerjepa,akbar2026weaksigreg}. Our method
extends this characteristic-function mechanism to two-sample
matching between arbitrary endpoint distributions within a single
invertible map.

\section{Method}
\label{sec:method}

Let $P_{\mathcal X}$ denote the data distribution and
$P_{\mathcal Z}$ a prescribed prior available only through samples. We
assume both distributions are defined on $\mathbb R^d$ and independent samples
$x\sim P_{\mathcal X}$ and $z\sim P_{\mathcal Z}$ are available; no
density evaluation is required. We seek an invertible map
$F_\theta:\mathbb R^d\rightarrow\mathbb R^d$, with inverse
$G_\theta=F_\theta^{-1}$, such that $
    (F_\theta)_{\#}P_{\mathcal X}=P_{\mathcal Z},
    (G_\theta)_{\#}P_{\mathcal Z}=P_{\mathcal X}$.
    
Exact consistency assumes that the endpoint distributions have equal
ambient dimension and admit an invertible transport representable by the
chosen architecture. Empirical collections are interpreted as samples
from underlying distributions rather than as finite discrete measures.

\subsection{Invertible Endpoint Map}
\label{sec:method-map}

The objective can be combined with any differentiable invertible
architecture. In our experiments, the map is induced by a neural ODE
\cite{chen2018neuralode},
\begin{equation}
    {d y_t}/{dt}=v_\theta(y_t,t),
    \qquad t\in[0,1],
    \label{eq:neural-ode}
\end{equation}
where $v_\theta$ is a neural vector field. Denoting its solution map from
time $s$ to $t$ by $\Phi_\theta^{s\rightarrow t}$, we define
\begin{equation}
    G_\theta=\Phi_\theta^{0\rightarrow1},
    \qquad
    F_\theta=\Phi_\theta^{1\rightarrow0}.
    \label{eq:forward-inverse}
\end{equation}
When the ODE has a unique solution, the two maps are inverses at the continuous level; fixed-step numerical integration produces approximate inverse maps in implementation. Unlike likelihood-trained continuous normalizing flows \cite{grathwohl2019ffjord}, the dynamics are used only as an invertible parameterization: training does not evaluate a base density or a change-of-variables divergence.

\subsection{Projected Characteristic-Function Matching}
\label{sec:method-pcf}

Our discrepancy adapts the statistical mechanism used by SIGReg in
LeJEPA \cite{balestriero2025lejepa}. SIGReg applies an
Epps--Pulley-style empirical-characteristic-function statistic to random
one-dimensional projections and compares them with an analytical
Gaussian target \cite{epps1983test}. We instead use a two-sample version
in which both projected laws are represented empirically, allowing an
arbitrary sample-only target.

Let $U=\{u_i\}_{i=1}^{B}$ and $V=\{v_i\}_{i=1}^{B}$ be minibatches in
$\mathbb R^d$. For a unit direction
$a_m\in\mathbb S^{d-1}$ and frequency $\omega_k$, define
\begin{equation}
    \widehat{\varphi}_{U}^{(m)}(\omega_k)
    =
    ({1}/{B})
    \textstyle\sum_{i=1}^{B}
    \exp\!\left(i\omega_k a_m^\top u_i\right),
    \label{eq:empirical-cf}
\end{equation}
and analogously $\widehat{\varphi}_{V}^{(m)}$. Given nonnegative
frequency weights satisfying $\sum_{k=1}^{K}w_k=1$, the discrepancy on
projection $a_m$ is
\begin{equation}
    r_m(U,V)
    =
    \frac{1}{4+\varepsilon}
    \sum_{k=1}^{K}w_k
    \left|
        \widehat{\varphi}_{U}^{(m)}(\omega_k)
        -
        \widehat{\varphi}_{V}^{(m)}(\omega_k)
    \right|^2 .
    \label{eq:pcf-slice}
\end{equation}
Since characteristic functions have modulus at most one,
$0\leq r_m<1$. We aggregate the projection discrepancies
\begin{equation}
    \widehat{\mathcal D}_{\mathrm{PCF}}(U,V)
    =
    ({1}/{M})\textstyle\sum_{m=1}^{M}
    \rho\!\left(r_m(U,V)\right),
    \label{eq:pcf-discrepancy}
\end{equation}
with
\begin{equation}
    \rho(r)
    =
    \log(1+\varepsilon)-\log(1-r+\varepsilon).
    \label{eq:pcf-contrast}
\end{equation}
The shifted contrast satisfies $\rho(0)=0$ and places greater emphasis
on poorly matched projections without changing the exact
zero-discrepancy solutions.

The complex modulus is evaluated through its sine and cosine components.
Projection directions are resampled during training, following SIGReg,
so that successive updates examine different one-dimensional marginals
\cite{balestriero2025lejepa}. Computing the projections and Fourier
features costs
$\mathcal O(BM(d+K))$, which is linear in the minibatch size for fixed
$M$, $K$, and $d$, whereas full kernel MMD requires pairwise
$\mathcal O(B^2d)$ evaluations.

A weighted squared characteristic-function distance can itself be viewed
as a spectral kernel discrepancy
\cite{ansari2020characteristic,zimmermann2025kerjepa}. PCF differs from
Direct-CF matching through its structured direction--frequency grid and
the projection-wise nonlinear aggregation in
Eq.~\eqref{eq:pcf-discrepancy}; these design choices are examined in the
ablation study.

\subsection{Bidirectional Endpoint Objective}
\label{sec:method-endpoint}

At each update, we independently sample both distributions, $x_1,\ldots,x_B\sim P_{\mathcal X}$,
    $z_1,\ldots,z_B\sim P_{\mathcal Z}$,
without defining correspondences between minibatches. The PCF
objective is
\begin{align}
    \widehat{\mathcal L}_{\mathrm{PCF}}(\theta)
    ={}&
    \frac{1}{2}
    \widehat{\mathcal D}_{\mathrm{PCF}}
    \left(
        \{F_\theta(x_i)\}_{i=1}^{B},
        \{z_i\}_{i=1}^{B}
    \right)
    \nonumber\\
    &+
    \frac{1}{2}
    \widehat{\mathcal D}_{\mathrm{PCF}}
    \left(
        \{G_\theta(z_i)\}_{i=1}^{B},
        \{x_i\}_{i=1}^{B}
    \right).
    \label{eq:pcf-loss}
\end{align}
The first term aligns encoded data with the prescribed prior, while the
second directly trains the generative direction. This symmetric
endpoint structure parallels sample-based bidirectional transport
objectives such as SyMOT-Flow, replacing pairwise MMD by the proposed
projected characteristic-function discrepancy
\cite{xiong2023symotflow}. Although either endpoint equality implies the
other for an exact invertible map at the population level, using both
terms provides direct finite-sample gradients in both directions.

\subsection{Transport-Regularized Matching}
\label{sec:method-transport}

Endpoint distributions do not identify a unique transport map. When two endpoints are represented in a common standardized Euclidean
coordinate system, we optionally use the symmetric displacement cost
\cite{xiong2023symotflow}
\begin{equation}
    \widehat{\mathcal R}_{\mathrm{TC}}(\theta)
    =
    \frac{1}{2dB}
    \sum_{i=1}^{B}
    \left[
        \left\|G_\theta(z_i)-z_i\right\|_2^2
        +
        \left\|F_\theta(x_i)-x_i\right\|_2^2
    \right].
    \label{eq:tc}
\end{equation}
The regularized training objective is
\begin{equation}
    \widehat{\mathcal L}_{\mathrm{PCF+TC}}(\theta)
    =
    \widehat{\mathcal L}_{\mathrm{PCF}}(\theta)
    +
    \lambda\widehat{\mathcal R}_{\mathrm{TC}}(\theta).
    \label{eq:pcf-tc}
\end{equation}
The PCF term enforces the endpoint laws, whereas the transport term
prefers lower-displacement maps among approximately valid solutions.
Since a large $\lambda$ can favour an identity-like map at the expense
of endpoint alignment, we treat transport regularization as an optional
weak geometric preference.

\subsection{Endpoint Consistency and Map Ambiguity}
\label{sec:method-theory}

Let $\mathcal D_{\mathrm{PCF}}^\infty$ denote the idealized population
discrepancy that compares projected characteristic functions over all unit
directions and all real frequencies. Since characteristic functions uniquely
identify one-dimensional distributions and the Cramér--Wold theorem identifies
a multivariate distribution from all of its one-dimensional projections
\cite{cramer1936some}, we obtain
\begin{equation}
    \mathcal D_{\mathrm{PCF}}^\infty(P,Q)=0
    \quad\Longleftrightarrow\quad
    P=Q.
    \label{eq:pcf-definite}
\end{equation}
Consequently, for an invertible $F_\theta$ with inverse $G_\theta$,
\begin{equation}
    \mathcal L_{\mathrm{PCF}}^\infty(\theta)=0
    \;\Longleftrightarrow\;
    (F_\theta)_{\#}P_{\mathcal X}=P_{\mathcal Z}
    \;\Longleftrightarrow\;
    (G_\theta)_{\#}P_{\mathcal Z}=P_{\mathcal X}.
    \label{eq:endpoint-consistency}
\end{equation}
Thus, endpoint consistency is a property of the idealized all-direction,
all-frequency objective. The implemented objective in
Eq.~\eqref{eq:pcf-loss} uses finitely many directions and frequencies and
finite minibatches; it is therefore a practical surrogate and is not claimed
to be characteristic for fixed $M$, $K$, and $B$. Moreover, the squared
empirical characteristic-function difference is a finite-sample
V-statistic whose expectation contains distribution-dependent terms of order
$\mathcal O(B^{-1})$. This bias vanishes as $B$ increases for the
corresponding finite-grid population quantity.

This result determines the endpoint distributions, but not a unique
sample correspondence. For example, if $H$ is an invertible
prior-preserving transformation,
$H_{\#}P_{\mathcal Z}=P_{\mathcal Z}$, then
$H\circ F_\theta$ satisfies the same endpoint condition as
$F_\theta$. This ambiguity is unavoidable without paired observations
or additional structural assumptions. The optional transport cost
prefers lower-displacement solutions, but does not guarantee recovery of
a unique ground-truth or optimal-transport map.

\subsection{Optimization}
\label{sec:method-optimization}

At each iteration, the forward and inverse maps are evaluated on
independent minibatches from $P_{\mathcal X}$ and $P_{\mathcal Z}$, and
the parameters are optimized using
Eq.~\eqref{eq:pcf-loss}, optionally augmented with the transport cost in
Eq.~\eqref{eq:pcf-tc}. Projection directions are resampled at every
iteration to increase projection coverage.

\section{Experiments}
\label{sec:experiments}

\subsection{Experimental Setup}
\label{sec:experimental-setup}

All methods are implemented as neural ODEs on \(t\in[0,1]\). Within each experiment, we use identical data splits, preprocessing, flow architecture, ODE solver, initialization, minibatch streams, and source and target sample exposure. Source and target minibatches are sampled independently. All standardization statistics are computed from the training split.

Unless otherwise stated, the low-dimensional experiments use an MLP with four hidden layers of width 128 and SiLU activations that takes the state and time as input. The ODE is integrated in both directions with the explicit midpoint rule, using 12 steps for training and 30 for evaluation (60 in the ablations). Training uses Adam with learning rate \(5\times10^{-4}\) and gradient clipping at norm 5. In the controlled benchmark, each method receives 512,000 training draws per endpoint, i.e., 500 updates at batch size 1,024 (methods with smaller batches take proportionally more updates); the ablations use 256 updates, except the architecture and ODE studies (128 updates) and the batch study (65,536 draws) in Table~\ref{tab:trained-robustness}. The high-dimensional experiments instead use the convolutional CNF and task-specific sample budgets.

The default PCF configuration uses 32 projection directions and 12 logarithmically spaced frequencies from \(0.25\) to \(8\) with weights \(w_k\propto\exp(-\omega_k^2/32)\), directions resampled at every update, and \(\epsilon=10^{-6}\); the high-dimensional experiments use 128 directions and 8 frequencies. Direct-CF uses 384 random Fourier features with independent uniform directions and frequencies drawn with probability \(w_k\). MMD uses 5 Gaussian kernels with bandwidths \(\{0.25,0.5,1,2,4\}\) times the median distance. Endpoint MMD is evaluated on held-out samples with the MMD baseline's training kernel; two independent samples of the true distributions score \(0.37\) (\(X\)) and \(0.45\) (\(Z\)). The C2ST classifier is a two-layer MLP of width 64 trained on a 70/30 split. The controlled benchmark and transport-cost sweep use five matched seeds; the ablation and high-dimensional studies use three because of computational constraints. We report sample average \(\pm\) std. deviation across seeds.

\subsection{Illustrative Examples}
\label{sec:illustrative-examples}

Figure~\ref{fig:illustrative-generation} shows bidirectional endpoint
alignment during training. PCF progressively matches both the data and
prescribed-prior distributions. In contrast, the Gaussian-prior CNF can
improve the data endpoint, but its encoded samples follow the imposed
Gaussian rather than the prescribed prior.

\newcommand{\trainpanel}[2][0.255\linewidth]{%
  \includegraphics[width=#1]{\detokenize{Figures/#2}}%
}

\newcommand{\refpanel}[1]{%
  \includegraphics[width=0.405\linewidth]{\detokenize{Figures/#1}}%
}

\newcommand{\trainstripcrop}[2]{%
  \includegraphics[
    width=0.255\linewidth,
    trim=#2,
    clip
  ]{\detokenize{Figures/#1}}%
}

\newcommand{\refstripcrop}[2]{%
  \includegraphics[
    width=0.405\linewidth,
    trim=#2,
    clip
  ]{\detokenize{Figures/#1}}%
}

\begin{figure*}[t]
\centering
\begingroup
\setlength{\tabcolsep}{1pt}
\renewcommand{\arraystretch}{0.92}

\begin{minipage}[t]{0.325\textwidth}
\centering
\textbf{Two moons}\\

\refpanel{media_images_viz_mle_gauss_2000_4cd35805eb5f5fa46f40_real_data.png}
\hspace{0.02\linewidth}
\refstripcrop
  {media_images_viz_mle_gauss_2000_4cd35805eb5f5fa46f40_prior.png}
  {0 0 0 0}
\\[-0.35em]
{\scriptsize Real data \hspace{0.25\linewidth} Prescribed prior}

\vspace{0.25em}
\begin{tabular}{@{}lccc@{}}
& \scriptsize 250
& \scriptsize 1,000
& \scriptsize 2,000 \\

\rotatebox{90}{\scriptsize PCF gen.}
&
\trainpanel{media_images_viz_sigflow_2000_77200e9fa184679fbf3d_generated.png}
&
\trainpanel{media_images_viz_sigflow_10000_c5cbd8c9b264e95bf7ab_generated.png}
&
\trainpanel{media_images_viz_sigflow_20000_58a8f75cdfcb8eccc054_generated.png}
\\[-0.25em]

\rotatebox{90}{\scriptsize PCF enc.}
&
\trainpanel{media_images_viz_sigflow_2000_77200e9fa184679fbf3d_encoded.png}
&
\trainpanel{media_images_viz_sigflow_10000_c5cbd8c9b264e95bf7ab_encoded.png}
&
\raisebox{0.8ex}{\trainpanel[0.21\linewidth]{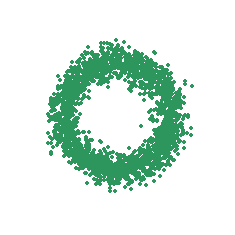}}

\\[-0.25em]

\rotatebox{90}{\scriptsize G-CNF gen.}
&
\trainpanel{media_images_viz_mle_gauss_2000_4cd35805eb5f5fa46f40_generated.png}
&
\trainpanel{media_images_viz_mle_gauss_10000_00a3e1afd942b593691e_generated.png}
&
\trainpanel{media_images_viz_mle_gauss_20000_201f70b4ae7c47af0c37_generated.png}
\\[-0.25em]

\rotatebox{90}{\scriptsize G-CNF enc.}
&
\trainpanel{media_images_viz_mle_gauss_2000_4cd35805eb5f5fa46f40_encoded.png}
&
\trainpanel{media_images_viz_mle_gauss_10000_00a3e1afd942b593691e_encoded.png}
&
\trainpanel{media_images_viz_mle_gauss_20000_201f70b4ae7c47af0c37_encoded.png}
\end{tabular}
\end{minipage}
\hfill
\begin{minipage}[t]{0.325\textwidth}
\centering
\textbf{Two spirals}\\

\refpanel{two_spirals__gaussian_cnf__step_001000_real_data.png}
\hspace{0.02\linewidth}
\refpanel{two_spirals__gaussian_cnf__step_001000_prior.png}
\\[-0.35em]
{\scriptsize Real data \hspace{0.25\linewidth} Prescribed prior}

\vspace{0.25em}
\begin{tabular}{@{}lccc@{}}
& \scriptsize 250
& \scriptsize 1,000
& \scriptsize 2,000 \\

\rotatebox{90}{\scriptsize PCF gen.}
&
\trainpanel{two_spirals__pcf__step_000250_generated.png}
&
\trainpanel{two_spirals__pcf__step_001000_generated.png}
&
\trainpanel{two_spirals__pcf__step_002000_generated.png}
\\[-0.25em]

\rotatebox{90}{\scriptsize PCF enc.}
&
\trainpanel{two_spirals__pcf__step_000250_generated_encoded.png}
&
\trainpanel{two_spirals__pcf__step_001000_encoded.png}
&
\trainpanel{two_spirals__pcf__step_002000_encoded.png}
\\[-0.25em]

\rotatebox{90}{\scriptsize G-CNF gen.}
&
\trainpanel{two_spirals__gaussian_cnf__step_000250_generated.png}
&
\trainpanel{two_spirals__gaussian_cnf__step_001000_generated.png}
&
\trainpanel{two_spirals__gaussian_cnf__step_002000_generated.png}
\\[-0.25em]

\rotatebox{90}{\scriptsize G-CNF enc.}
&
\trainpanel{two_spirals__gaussian_cnf__step_000250_encoded.png}
&
\trainpanel{two_spirals__gaussian_cnf__step_001000_encoded.png}
&
\trainpanel{two_spirals__gaussian_cnf__step_002000_encoded.png}
\end{tabular}
\end{minipage}
\hfill
\begin{minipage}[t]{0.325\textwidth}
\centering
\textbf{Cellular checkerboard}\\

\refpanel{cellular_checkerboard__gaussian_cnf__step_000250_real_data.png}
\hspace{0.02\linewidth}
\refpanel{cellular_checkerboard__gaussian_cnf__step_000250_prior.png}
\\[-0.35em]
{\scriptsize Real data \hspace{0.25\linewidth} Prescribed prior}

\vspace{0.25em}
\begin{tabular}{@{}lccc@{}}
& \scriptsize 250
& \scriptsize 1,000
& \scriptsize 2,000 \\

\rotatebox{90}{\scriptsize PCF gen.}
&
\trainpanel{cellular_checkerboard__pcf__step_000250_generated.png}
&
\trainpanel{cellular_checkerboard__pcf__step_001000_generated.png}
&
\trainpanel{cellular_checkerboard__pcf__step_002000_generated.png}
\\[-0.25em]

\rotatebox{90}{\scriptsize PCF enc.}
&
\trainpanel{cellular_checkerboard__pcf__step_000250_encoded.png}
&
\trainpanel{cellular_checkerboard__pcf__step_001000_encoded.png}
&
\trainpanel{cellular_checkerboard__pcf__step_002000_encoded.png}
\\[-0.25em]

\rotatebox{90}{\scriptsize G-CNF gen.}
&
\trainpanel{cellular_checkerboard__gaussian_cnf__step_000250_generated.png}
&
\trainpanel{cellular_checkerboard__gaussian_cnf__step_001000_generated.png}
&
\trainpanel{cellular_checkerboard__gaussian_cnf__step_002000_generated.png}
\\[-0.25em]

\rotatebox{90}{\scriptsize G-CNF enc.}
&
\trainpanel{cellular_checkerboard__gaussian_cnf__step_000250_encoded.png}
&
\trainpanel{cellular_checkerboard__gaussian_cnf__step_001000_encoded.png}
&
\trainpanel{cellular_checkerboard__gaussian_cnf__step_002000_encoded.png}
\end{tabular}
\end{minipage}

\endgroup

\caption{
Bidirectional endpoint alignment on three structured sample-only priors.
Columns show the endpoint distributions and generated (\emph{gen.}) and encoded (\emph{enc.}) samples after 250, 1,000, and 2,000 updates. PCF aligns both endpoints, whereas G-CNF
encodes samples to its imposed Gaussian base.
}
\label{fig:illustrative-generation}
\end{figure*}

In Fig.~\ref{fig:illustrative-paths}, PCF learns coherent bidirectional
maps from independent endpoint samples, whereas independent-coupling Flow
Matching produces substantial trajectory crossing.

\newcommand{\smalltrajectorypanel}[1]{%
    \includegraphics[width=0.47\linewidth]{Figures/#1}%
}

\begin{figure}[t]
\centering

\begin{minipage}[t]{0.49\columnwidth}
\centering
\textbf{\scriptsize Eight Gaussians}\\

\begin{tabular}{@{}cc@{}}
\smalltrajectorypanel{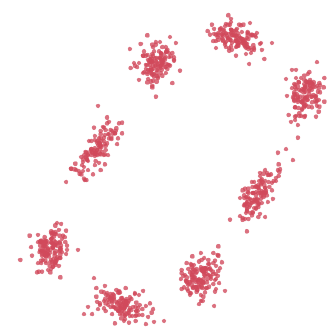}
&
\smalltrajectorypanel{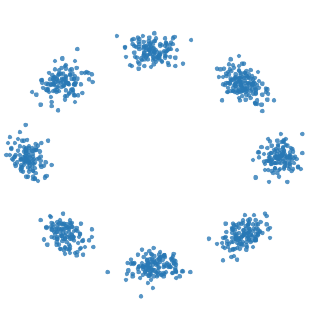}
\\[-0.35em]
{\tiny Real data}
&
{\tiny Prescribed prior}
\\[0.05em]

\smalltrajectorypanel{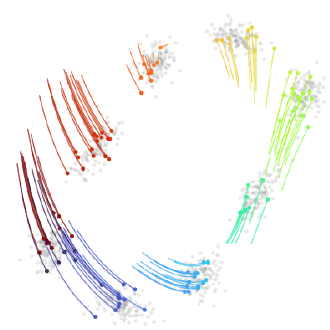}
&
\smalltrajectorypanel{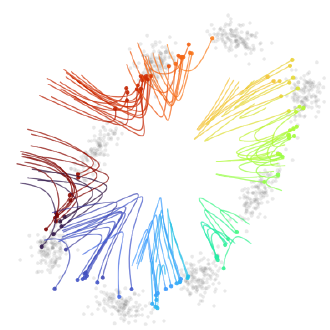}
\\[-0.35em]
{\tiny PCF: prior $\rightarrow$ data}
&
{\tiny FM: prior $\rightarrow$ data}
\\[0.05em]

\smalltrajectorypanel{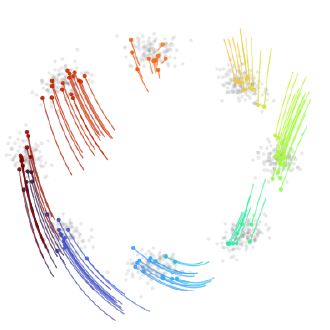}
&
\smalltrajectorypanel{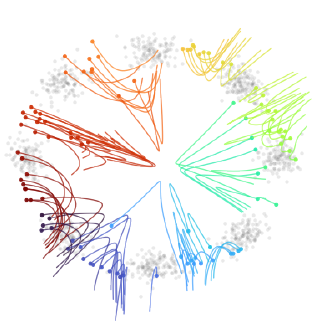}
\\[-0.35em]
{\tiny PCF: data $\rightarrow$ prior}
&
{\tiny FM: data $\rightarrow$ prior}
\end{tabular}
\end{minipage}
\hfill
\begin{minipage}[t]{0.49\columnwidth}
\centering
\textbf{\scriptsize Anisotropic Gaussian}\\

\begin{tabular}{@{}cc@{}}
\smalltrajectorypanel{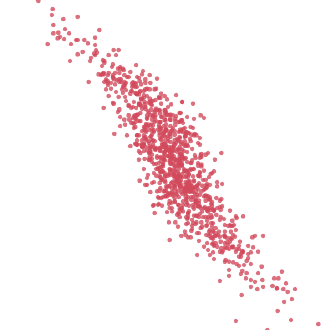}
&
\smalltrajectorypanel{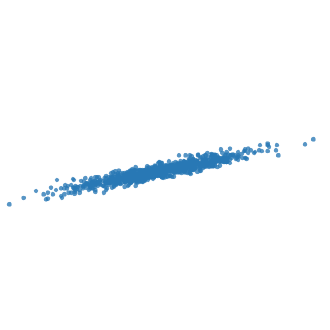}
\\[-0.35em]
{\tiny Real data}
&
{\tiny Prescribed prior}
\\[0.05em]

\smalltrajectorypanel{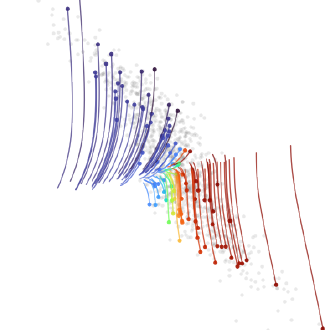}
&
\smalltrajectorypanel{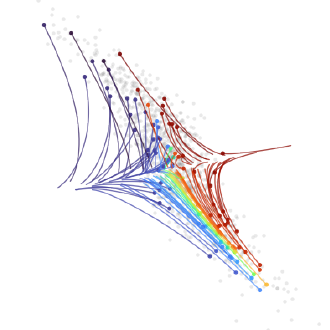}
\\[-0.35em]
{\tiny PCF: prior $\rightarrow$ data}
&
{\tiny FM: prior $\rightarrow$ data}
\\[0.05em]

\smalltrajectorypanel{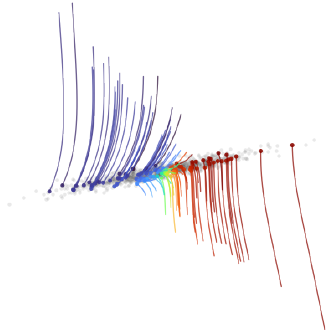}
&
\smalltrajectorypanel{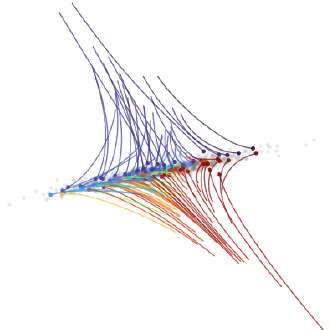}
\\[-0.35em]
{\tiny PCF: data $\rightarrow$ prior}
&
{\tiny FM: data $\rightarrow$ prior}
\end{tabular}
\end{minipage}

\caption{
Bidirectional trajectories on multimodal and anisotropic tasks.
PCF uses unpaired endpoint samples, while independent-coupling Flow Matching
exhibits greater trajectory crossing.
}
\label{fig:illustrative-paths}
\end{figure}

\subsection{Controlled Sample-Only Transport}
\label{sec:experiments-controlled}

We first evaluate bidirectional transport on four controlled worlds:
Gaussian, multimodal Eight Gaussians, anisotropic Gaussian, and nested
rings similar to \cite{xiong2023symotflow}. In each world, the data distribution is obtained from the prescribed prior through a known invertible teacher. Paired samples are retained only
to evaluate teacher recovery; training uses independent, unpaired
minibatches from the two endpoint laws.

To isolate the effect of the endpoint discrepancy, PCF, MMD
\cite{gretton2012kernel}, Direct-CF, sliced Wasserstein
\cite{rabin2011wasserstein,bonneel2015sliced}, energy distance, and
Sinkhorn divergence \cite{cuturi2013sinkhorn,genevay2018learning} use the
same symmetric bidirectional objective. We additionally compare with Flow
Matching using independent and minibatch-OT couplings
\cite{lipman2023flowmatching,pooladian2023multisample}, and with
likelihood-trained CNFs \cite{chen2018neuralode,grathwohl2019ffjord}
using a Gaussian, a fitted full-covariance GMM, or a separately fitted flow
as the prior density.

We report data-endpoint alignment
$\mathrm{MMD}_{X}$, prior-endpoint alignment $\mathrm{MMD}_{Z}$, and the
average C2ST gap from chance. Teacher RMSE evaluates recovery of planted
map, while TC measures average displacement on forward and inverse
transports. TC is a geometric diagnostic and must be interpreted jointly
with endpoint alignment. Table~\ref{tab:main-synthetic} reports macro
averages over four worlds and five matched seeds.

\begin{table*}[t]
\centering
\small
\setlength{\tabcolsep}{4pt}
\caption{
Controlled transport benchmark. Endpoint MMD values are scaled by $10^3$;
TC measures displacement rather than endpoint error. Values are mean
$\pm$ sample standard deviation.
}
\label{tab:main-synthetic}
\resizebox{\linewidth}{!}{
\begin{tabular}{lccccc}
\toprule
Method
& $\mathrm{MMD}_{X}\downarrow$
& $\mathrm{MMD}_{Z}\downarrow$
& $\mathrm{C2ST\text{-}Gap}_{\mathrm{avg}}\downarrow$
& $\mathrm{Teacher\ RMSE}\downarrow$
& $\mathrm{TC}$ \\
\midrule
\multicolumn{6}{l}{\emph{Sample-based endpoint matching}} \\

PCF
& $\mathbf{0.6487 \pm 0.1523}$
& $\underline{0.5675 \pm 0.1093}$
& $0.0116 \pm 0.0027$
& $0.257 \pm 0.008$
& $0.597 \pm 0.011$ \\

PCF+TC
& $\underline{0.6569 \pm 0.1496}$
& $\mathbf{0.5672 \pm 0.1059}$
& $0.0122 \pm 0.0032$
& $\mathbf{0.246 \pm 0.006}$
& $0.581 \pm 0.011$ \\

MMD
& $0.6659 \pm 0.1537$
& $0.5757 \pm 0.1140$
& $\underline{0.0103 \pm 0.0017}$
& $0.258 \pm 0.002$
& $0.598 \pm 0.013$ \\

MMD+TC
& $0.6768 \pm 0.1505$
& $0.5775 \pm 0.1088$
& $\mathbf{0.0098 \pm 0.0017}$
& $\underline{0.248 \pm 0.003}$
& $0.581 \pm 0.012$ \\

Direct-CF
& $0.6664 \pm 0.1653$
& $0.5818 \pm 0.1124$
& $0.0132 \pm 0.0037$
& $0.259 \pm 0.003$
& $0.598 \pm 0.013$ \\

Sliced-Wasserstein
& $0.9046 \pm 0.2494$
& $0.6515 \pm 0.1502$
& $0.0251 \pm 0.0066$
& $0.262 \pm 0.006$
& $0.606 \pm 0.025$ \\

Energy-Distance
& $0.6833 \pm 0.1805$
& $0.6061 \pm 0.1553$
& $0.0129 \pm 0.0051$
& $0.262 \pm 0.004$
& $0.606 \pm 0.017$ \\

Sinkhorn-Divergence
& $1.0681 \pm 0.3209$
& $0.7623 \pm 0.1685$
& $0.0181 \pm 0.0046$
& $0.259 \pm 0.011$
& $0.595 \pm 0.022$ \\

\midrule
\multicolumn{6}{l}{\emph{Path-based flow training}} \\

FM--Independent Coupling
& $2.5732 \pm 0.7546$
& $1.9559 \pm 0.4955$
& $0.0767 \pm 0.0129$
& $0.590 \pm 0.010$
& $0.632 \pm 0.026$ \\

FM--Minibatch OT Coupling
& $1.7554 \pm 0.1178$
& $1.2446 \pm 0.1373$
& $0.1055 \pm 0.0045$
& $0.378 \pm 0.003$
& $0.529 \pm 0.014$ \\

\midrule
\multicolumn{6}{l}{\emph{Likelihood-based flow training}} \\

Gaussian Prior
& $50.6638 \pm 5.0390$
& $31.1678 \pm 0.9266$
& $0.2278 \pm 0.0077$
& $0.615 \pm 0.029$
& $0.289 \pm 0.024$ \\

64-Component Full-Covariance GMM Prior
& $1.1728 \pm 0.4226$
& $1.0743 \pm 0.3688$
& $0.0127 \pm 0.0029$
& $0.248 \pm 0.057$
& $0.623 \pm 0.009$ \\

Flow-Fitted Prior
& $0.7228 \pm 0.1712$
& $0.6720 \pm 0.1242$
& $0.0112 \pm 0.0046$
& $0.328 \pm 0.108$
& $0.603 \pm 0.009$ \\

\bottomrule
\end{tabular}
}
\end{table*}

Table~\ref{tab:main-synthetic} shows that direct endpoint objectives obtain
the lowest mean bidirectional endpoint errors. PCF has the lowest mean
$\mathrm{MMD}_{X}$, PCF+TC has the lowest mean $\mathrm{MMD}_{Z}$, and the
MMD variants obtain slightly lower mean C2ST gaps. The differences between
PCF and MMD are small relative to the across-seed variation, so we regard
their controlled-benchmark endpoint performance as comparable. Direct-CF and
energy distance also remain competitive, whereas sliced Wasserstein and
Sinkhorn are less accurate on these tasks.

Flow Matching produces lower-displacement maps with minibatch-OT coupling but
weaker endpoint alignment. Likelihood baselines improve when the Gaussian
prior is replaced by a fitted GMM or flow, but they require an additional
density-estimation stage. Overall, directly optimizing the endpoint laws is
more accurate than the path-based and likelihood-based alternatives on these
controlled tasks.

\subsection{Endpoint Matching and Transport Regularization}
\label{sec:experiments-regularization}

Endpoint alignment does not identify a unique invertible map. We therefore
study whether the optional transport term in Eq.~\eqref{eq:tc} can select
lower-displacement solutions without substantially degrading the endpoint
laws. We sweep $\lambda$ from \(0\) to \(1\) for PCF and MMD over the four
controlled worlds and five matched seeds. For each objective, validation
selection chooses the positive weight with the lowest TC whose average
endpoint MMD is at most \(5\%\) above its unregularized value.

The selected PCF weight, \(\lambda=3\times10^{-4}\), reduces TC by \(2.92\%\)
while increasing average endpoint MMD by only \(0.83\%\). For MMD,
\(\lambda=10^{-3}\) reduces TC by \(2.95\%\) with a \(1.05\%\) increase in
endpoint MMD. In both cases, the C2ST gap remains essentially unchanged.
These operating points correspond to the PCF+TC and MMD+TC models reported
in Table~\ref{tab:main-synthetic}. The absolute values of \(\lambda\) are not
directly comparable since discrepancies have different
scales.

\begin{figure}[t]
    \centering
    \begin{subfigure}[t]{0.48\columnwidth}
        \centering
        \includegraphics[width=\linewidth]
        {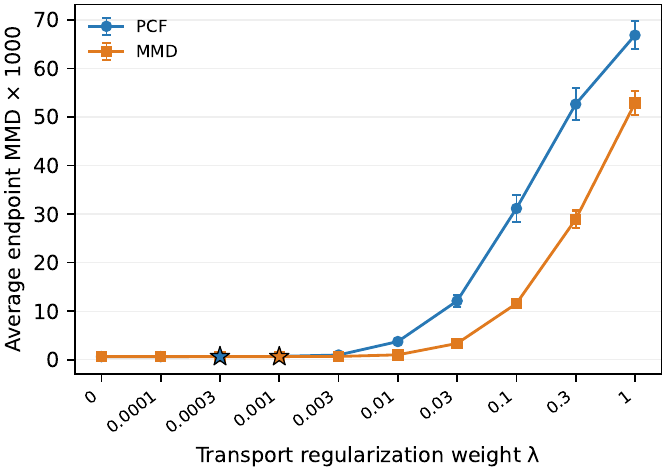}
        \caption{Endpoint error.}
        \label{fig:tc-endpoint-lambda}
    \end{subfigure}
    \hfill
    \begin{subfigure}[t]{0.48\columnwidth}
        \centering
        \includegraphics[width=\linewidth]
        {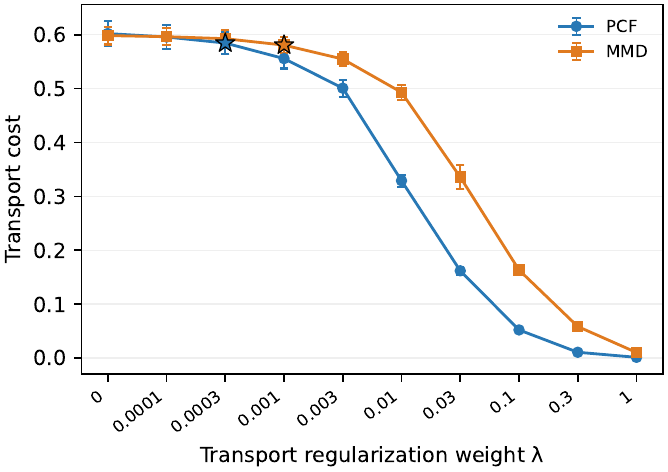}
        \caption{Transport cost.}
        \label{fig:tc-cost-lambda}
    \end{subfigure}

    \vspace{0.5em}

    \begin{subfigure}[t]{0.48\columnwidth}
        \centering
        \includegraphics[width=\linewidth]
        {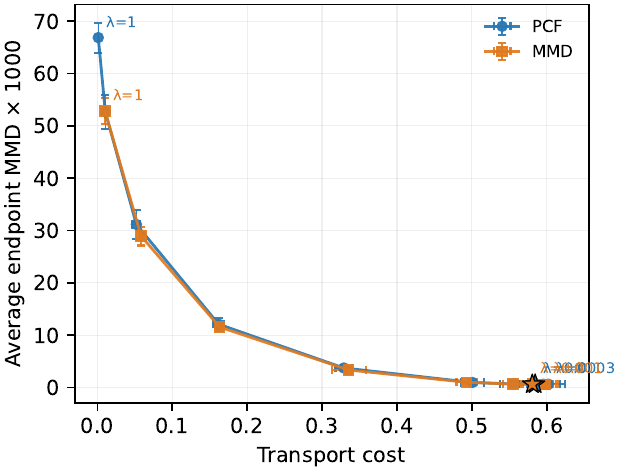}
        \caption{Endpoint--cost trade-off.}
        \label{fig:tc-pareto-mmd}
    \end{subfigure}
    \hfill
    \begin{subfigure}[t]{0.48\columnwidth}
        \centering
        \includegraphics[width=\linewidth]
        {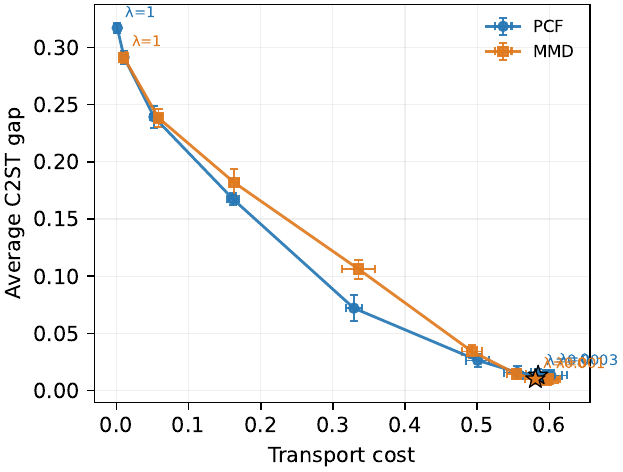}
        \caption{C2ST--cost trade-off.}
        \label{fig:tc-pareto-c2st}
    \end{subfigure}

    \caption{
    Transport-cost sweep. Stars denote the validation-selected operating points. Weak regularization reduces displacement while preserving endpoint alignment; large weights sacrifice alignment.
    }
    \label{fig:transport-regularization}
\end{figure}

In Fig.~\ref{fig:transport-regularization}, TC is useful as a weak
geometric preference rather than as the dominant training objective. Small
weights select simpler maps retaining the endpoint accuracy;
strong regularization biases the flow toward
low-displacement transformations that no longer connect the prescribed laws.

\subsection{Ablation and Scalability Analysis}
\label{sec:experiments-ablation}

We analyze PCF's objective, representation, sample efficiency, and scaling.
The main ablations use the Gaussian, Eight Gaussians, anisotropic Gaussian,
and nested rings worlds. Frequency, projection, architecture, and ODE studies
use Eight Gaussians and anisotropic Gaussian; batch scaling uses Gaussian and
Eight Gaussians; and dimension scaling uses high-dimensional Gaussian and
multimodal transports. All configurations use three matched seeds. Endpoint
MMD is averaged over both directions and multiplied by \(10^3\).

\paragraph{Objective and representation.}
We vary the relative weight of the two endpoint losses and compare the default
PCF representation with alternative frequency weights, grids, aggregation
rules, and Fourier features. Table~\ref{tab:pcf-design-ablation} shows that
bidirectional supervision gives the most reliable practical optimization,
while the balance between the two losses has a smaller effect for
$\alpha\in\{0.25,0.5,0.75\}$. Multiscale complex features are important.
Uniform frequency weighting slightly improves the reported means, whereas
the projection-wise contrast \(\rho\) has a comparatively small effect.

\begin{table}[t]
\centering
\small
\setlength{\tabcolsep}{3.2pt}
\caption{
Objective and characteristic-feature ablations. Endpoint MMD is scaled by
$10^3$; comparisons are within blocks.
}
\label{tab:pcf-design-ablation}
\resizebox{\columnwidth}{!}{
\begin{tabular}{lcccc}
\toprule
Configuration
& Endpoint MMD $\downarrow$
& C2ST gap $\downarrow$
& Teacher RMSE $\downarrow$
& TC \\
\midrule
\multicolumn{5}{l}{\emph{Endpoint-objective ablation}} \\
\(\alpha=0.25\)
& \(0.674\) & \(0.01087\) & \(0.276\) & \(0.606\) \\
Symmetric PCF, \(\alpha=0.5\)
& \(0.642\) & \(0.01096\) & \(\mathbf{0.271}\) & \(0.601\) \\
\(\alpha=0.75\)
& \(\mathbf{0.640}\) & \(\mathbf{0.00897}\) & \(0.274\) & \(0.599\) \\
Encoding endpoint only, \(\alpha=1\)
& \(0.721\) & \(0.01288\) & \(0.278\) & \(0.577\) \\
Generation endpoint only, \(\alpha=0\)
& \(12.673\) & \(0.08062\) & \(2.334\) & \(33.292\) \\
\midrule
\multicolumn{5}{l}{\emph{Characteristic-feature ablation}} \\
Full PCF
& \(0.642\) & \(0.01096\) & \(0.271\) & \(0.601\) \\
Uniform frequency weights
& \(\mathbf{0.617}\) & \(\mathbf{0.00883}\) & \(\mathbf{0.270}\) & \(0.606\) \\
No projection-wise contrast
& \(0.648\) & \(0.01042\) & \(0.271\) & \(0.601\) \\
Direct-CF
& \(0.642\) & \(0.01073\) & \(0.273\) & \(0.597\) \\
No \(\rho\) (linear aggregation)
& \(0.646\) & \(0.01091\) & \(0.271\) & \(0.601\) \\
Single-scale PCF
& \(2.175\) & \(0.13645\) & \(0.356\) & \(0.592\) \\
Cosine-only PCF
& \(42.031\) & \(0.24896\) & \(0.482\) & \(0.409\) \\
\bottomrule
\end{tabular}
}
\end{table}

\paragraph{Frequency and projection design.}
We vary the frequency range, projection-resampling policy, and allocation of
features between directions and frequencies. Table~\ref{tab:pcf-design-robustness}
shows that broad multiscale grids are robust. Restricting the loss to low
frequencies degrades alignment, and high-frequency-only training performs
poorly (two of six runs diverged; as for \(\alpha=0\), with one of 12, we
average the seeds completed in all worlds). Projection resampling improves alignment, and
performance saturates at a moderate feature budget. Increasing the number of
directions cannot compensate for using too few frequencies.

\begin{table}[t]
\centering
\small
\setlength{\tabcolsep}{4pt}
\caption{
Frequency, projection-sampling, and feature-budget ablations.
Endpoint MMD is scaled by $10^3$.
}
\label{tab:pcf-design-robustness}
\resizebox{\columnwidth}{!}{
\begin{tabular}{lccc}
\toprule
Configuration
& Endpoint MMD $\downarrow$
& Teacher RMSE $\downarrow$
& Time/step (s) $\downarrow$ \\
\midrule
\multicolumn{4}{l}{\emph{Frequency design}} \\
Uniform weighting
& \(\mathbf{0.472}\) & \(\mathbf{0.0682}\) & \(0.0375\) \\
Default multiscale
& \(0.519\) & \(0.0720\) & \(0.0493\) \\
Low-frequency only
& \(2.831\) & \(0.1977\) & \(0.0479\) \\
High-frequency only
& \(56.599\) & \(1.3298\) & \(0.0507\) \\
\midrule
\multicolumn{4}{l}{\emph{Projection sampling}} \\
Per-update resampling
& \(\mathbf{0.519}\) & \(0.0720\) & \(0.0502\) \\
Per-epoch resampling
& \(0.570\) & \(\mathbf{0.0695}\) & \(\mathbf{0.0425}\) \\
Fixed projections
& \(0.596\) & \(0.0790\) & \(0.0523\) \\
\midrule
\multicolumn{4}{l}{\emph{Directions \(\times\) frequencies}} \\
\(8\times4\)
& \(0.797\) & \(0.0866\) & \(0.0357\) \\
\(16\times8\)
& \(0.582\) & \(0.0768\) & \(0.0319\) \\
\(32\times8\)
& \(0.514\) & \(\mathbf{0.0722}\) & \(0.0459\) \\
\(64\times8\)
& \(\mathbf{0.505}\) & \(0.0728\) & \(0.0422\) \\
\(64\times2\), 128 features
& \(21.285\) & \(0.3675\) & \(0.0437\) \\
\bottomrule
\end{tabular}
}
\end{table}

\paragraph{Sample efficiency and discrepancy scaling.}
We vary the number of unique endpoint samples while keeping the optimization
budget fixed, and separately benchmark discrepancy over dimensions
2--512 and batches 32--1,024. Table~\ref{tab:sample-and-scaling} shows that
PCF is strongest when the empirical libraries are small; it has lower error
than MMD in 13 of 15 paired (size, seed) runs. Its Fourier-feature
evaluation also matches Direct-CF in cost and avoids the pairwise memory growth
of MMD.

\begin{table}[t]
\centering
\small
\caption{
Sample-efficiency and discrepancy-scaling results. Endpoint MMD is scaled by
$10^3$; the discrepancy benchmark uses $d=512$ and $B=1{,}024$.
}
\label{tab:sample-and-scaling}

\begin{minipage}[t]{0.57\columnwidth}
\centering
\textbf{\scriptsize Sample efficiency}\\[2pt]
\resizebox{\linewidth}{!}{
\begin{tabular}{rccc}
\toprule
Samples
& PCF $\downarrow$
& Direct-CF $\downarrow$
& MMD $\downarrow$ \\
\midrule
\(256\)
& \(\mathbf{2.302 \pm 1.169}\) & \(2.667 \pm 1.814\) & \(2.599 \pm 1.565\) \\
\(1{,}024\)
& \(\mathbf{0.939 \pm 0.134}\) & \(0.945 \pm 0.168\) & \(1.001 \pm 0.141\) \\
\(4{,}096\)
& \(\mathbf{0.726 \pm 0.285}\) & \(0.776 \pm 0.297\) & \(0.805 \pm 0.288\) \\
\(8{,}192\)
& \(\mathbf{0.702 \pm 0.133}\) & \(0.721 \pm 0.081\) & \(0.738 \pm 0.043\) \\
\(16{,}384\)
& \(\mathbf{0.642 \pm 0.198}\) & \(\mathbf{0.642 \pm 0.143}\) & \(0.666 \pm 0.131\) \\
\bottomrule
\end{tabular}
}
\end{minipage}
\hfill
\begin{minipage}[t]{0.40\columnwidth}
\centering
\textbf{\scriptsize Discrepancy scaling}\\[2pt]
\resizebox{\linewidth}{!}{
\begin{tabular}{lcc}
\toprule
Method
& Time
& Memory \\
& (ms) $\downarrow$
& (MiB) $\downarrow$ \\
\midrule
PCF
& \(\mathbf{2.69}\)
& \(\mathbf{82.4}\) \\
Direct-CF
& \(2.70\)
& \(\mathbf{82.4}\) \\
MMD
& \(3.68\)
& \(117.4\) \\
\bottomrule
\end{tabular}
}
\end{minipage}
\end{table}

\paragraph{Architecture, ODE, batch, and dimension.}
We vary network capacity, training and evaluation ODE resolution, batch size,
and ambient dimension. Representative results appear in
Table~\ref{tab:trained-robustness}. PCF remains stable across architectures
and integration resolutions; coarse integration substantially reduces
runtime. Endpoint accuracy under batch and dimension scaling is mixed: MMD is
stronger in higher-dimensional tasks, PCF retains lower memory use
and competitive runtime. The main scalability advantage is therefore
feature-based memory efficiency rather than universal endpoint superiority.

\begin{table}[t]
\centering
\small
\caption{
Architecture and ODE robustness with batch and dimension scaling.
Paired entries are PCF/MMD; memory is reported in MiB.
}
\label{tab:trained-robustness}

\begin{minipage}[t]{0.49\columnwidth}
\centering
\textbf{\scriptsize Architecture and ODE}\\[2pt]
\resizebox{\linewidth}{!}{
\begin{tabular}{lccc}
\toprule
Setting
& PCF
& MMD
& Step ms \\
\midrule
\multicolumn{4}{l}{\emph{Architecture}} \\
\(64\times3\)
& \(4.062\) & \(\mathbf{3.681}\) & \(50.7/52.5\) \\
\(128\times4\)
& \(\mathbf{2.210}\) & \(2.688\) & \(49.6/54.1\) \\
\(256\times4\)
& \(0.752\) & \(\mathbf{0.747}\) & \(61.1/63.7\) \\
\midrule
\multicolumn{4}{l}{\emph{Training ODE steps}} \\
\(4\)
& \(\mathbf{2.152}\) & \(2.735\) & \(33.7/36.5\) \\
\(12\)
& \(\mathbf{2.210}\) & \(2.688\) & \(56.0/58.6\) \\
\(24\)
& \(\mathbf{2.228}\) & \(2.702\) & \(90.4/80.6\) \\
\midrule
\multicolumn{4}{l}{\emph{Evaluation ODE steps}} \\
\(12\)
& \(\mathbf{2.231}\) & \(2.703\) & -- \\
\(60\)
& \(\mathbf{2.210}\) & \(2.688\) & -- \\
\(120\)
& \(\mathbf{2.209}\) & \(2.687\) & -- \\
\bottomrule
\end{tabular}
}
\end{minipage}
\hfill
\begin{minipage}[t]{0.49\columnwidth}
\centering
\textbf{\scriptsize Batch and dimension}\\[2pt]
\resizebox{\linewidth}{!}{
\begin{tabular}{lcc}
\toprule
Setting
& Endpoint
& Memory \\
& PCF/MMD
& PCF/MMD \\
\midrule
\multicolumn{3}{l}{\emph{Batch size}} \\
\(B=32\)
& \(1.090/0.966\)
& \(6.8/6.7\) \\
\(B=128\)
& \(0.712/0.666\)
& \(25.8/26.3\) \\
\(B=512\)
& \(1.104/1.420\)
& \(101.9/122.9\) \\
\(B=1{,}024\)
& \(2.578/2.497\)
& \(203.5/297.2\) \\
\midrule
\multicolumn{3}{l}{\emph{Dimension}} \\
\(d=16\)
& \(3.100/2.271\)
& \(101.1/105.7\) \\
\(d=64\)
& \(2.128/2.075\)
& \(103.8/108.5\) \\
\(d=256\)
& \(1.738/1.578\)
& \(114.7/119.6\) \\
\(d=512\)
& \(1.672/1.476\)
& \(129.6/134.9\) \\
\multicolumn{3}{r}{\scriptsize Step ms at \(d=512\): \(52.2/58.1\)} \\
\bottomrule
\end{tabular}
}
\end{minipage}
\end{table}

\subsection{High-Dimensional Sample-Only Alignment}
\label{sec:experiments-high-dimensional}

We evaluate unpaired alignment on high-dimensional arrays, where the desired law is naturally represented by a finite collection of samples. MNIST and USPS
\cite{lecun1998gradient,hull1994database} are transported directly in
standardized $1\times28\times28$ pixel space ($d=784$). For BraTS
\cite{menze2015multimodal}, T1 and T2 slices are encoded by separately trained
autoencoders and transported in a $64\times24\times24$ latent space
($d=36{,}864$); reconstruction SSIMs are $0.971$ and $0.964$. All methods use
the same time-conditioned convolutional CNF and comparable endpoint exposure.

\begin{table}[!ht]
\centering
\small
\setlength{\tabcolsep}{3.2pt}
\caption{
High-dimensional alignment. Endpoint MMD is scaled by $10^3$;
times exclude the BraTS autoencoders.
}
\label{tab:high-dimensional}
\resizebox{\columnwidth}{!}{
\begin{tabular}{lcccc}
\toprule
Method
& Endpoint MMD $\downarrow$
& TC
& Time (s) $\downarrow$
& Memory (GiB) $\downarrow$ \\
\midrule
\multicolumn{5}{l}{\emph{MNIST--USPS, raw pixels}} \\

PCF
& $\underline{3.313 \pm 0.091}$
& $1.786 \pm 0.131$
& $\underline{566.82 \pm 0.42}$
& $\mathbf{40.90}$ \\

MMD
& $\mathbf{3.233 \pm 0.423}$
& $1.772 \pm 0.182$
& $\mathbf{565.72 \pm 0.37}$
& $\mathbf{40.90}$ \\

Direct-CF
& $4.298 \pm 0.660$
& $1.619 \pm 0.116$
& $572.74 \pm 0.65$
& $\mathbf{40.90}$ \\

\midrule
\multicolumn{5}{l}{\emph{BraTS T1--T2, autoencoder latents}} \\

PCF
& $22.553 \pm 0.424$
& $3.198 \pm 0.004$
& $\underline{146.46 \pm 1.28}$
& $\mathbf{4.10}$ \\

MMD
& $\mathbf{9.317 \pm 0.182}$
& $3.237 \pm 0.008$
& $\mathbf{144.13 \pm 0.52}$
& $\underline{4.11}$ \\

Direct-CF
& $\underline{19.749 \pm 0.497}$
& $3.202 \pm 0.008$
& $155.98 \pm 1.33$
& $4.13$ \\

\bottomrule
\end{tabular}
}
\end{table}

Table~\ref{tab:high-dimensional} shows that PCF is competitive with MMD and
outperforms Direct-CF on MNIST--USPS, while MMD is stronger on BraTS, where
batches of 64 samples meet \(d=36{,}864\). PCF pays off when the batch far
exceeds the number of directions: training a lighter MLP flow on raw
MNIST--USPS at batch size 16,384, a PCF step takes 70~ms and 4.0~GiB versus
700~ms and 31.8~GiB for MMD, which exceeds 46~GiB at 32,768 (PCF: 7.7~GiB).
We recommend 8--12 log-spaced frequencies in \([0.25,8]\) on standardized
data and \(\max(32,d/6)\) directions.
These unpaired experiments measure distributional rather than semantic
alignment; BraTS TC is descriptive because its latent spaces are trained
separately.

\section{Conclusion}

We introduced PCF, a density-free objective for learning invertible
transports between sample-only distributions by matching projected empirical
characteristic functions in both directions, without prior densities,
adversarial training, or source--target couplings. Across controlled and
raw-pixel experiments, PCF matches kernel MMD at linear cost, with
strong sample efficiency and stable bidirectional training, while an optional
transport regularizer selects lower-displacement maps. Future work should
explore adaptive projections and frequencies for high-dimensional structured
domains.

\bibliography{aaai2027}

\begin{thebibliography}{35}
\providecommand{\natexlab}[1]{#1}

\bibitem[{Akbar(2026)}]{akbar2026weaksigreg}
Akbar, H. 2026.
\newblock Weak-{SIGReg}: Covariance Regularization for Stable Deep Learning.
\newblock Accepted at the GRaM Workshop at ICLR 2026, arXiv:2603.05924.

\bibitem[{An, Hong, and Jeon(2023)}]{an2023mixturecw}
An, S.; Hong, S.; and Jeon, J.-J. 2023.
\newblock Balanced Marginal and Joint Distributional Learning via Mixture
  Cramer-Wold Distance.
\newblock arXiv:2312.03307.

\bibitem[{An and Jeon(2023)}]{an2023jointcw}
An, S.; and Jeon, J.-J. 2023.
\newblock Joint Distributional Learning via Cramer-Wold Distance.
\newblock arXiv:2310.16374.

\bibitem[{Ansari, Scarlett, and Soh(2020)}]{ansari2020characteristic}
Ansari, A.~F.; Scarlett, J.; and Soh, H. 2020.
\newblock A characteristic function approach to deep implicit generative
  modeling.
\newblock In \emph{Proc. IEEE/CVF conf. comp. vision patt. recog.}, 7478--7487.

\bibitem[{Balestriero and LeCun(2025)}]{balestriero2025lejepa}
Balestriero, R.; and LeCun, Y. 2025.
\newblock LeJEPA: Provable and Scalable Self-Supervised Learning Without the
  Heuristics.
\newblock arXiv:2511.08544.

\bibitem[{Bonneel et~al.(2015)Bonneel, Rabin, Peyr{\'e}, and
  Pfister}]{bonneel2015sliced}
Bonneel, N.; Rabin, J.; Peyr{\'e}, G.; and Pfister, H. 2015.
\newblock Sliced and Radon Wasserstein Barycenters of Measures.
\newblock \emph{J. Math. Imag. Vision}, 51(1): 22--45.

\bibitem[{Chao et~al.(2023)Chao, Sun, Hsu, Kira, and Lee}]{chao2023ebflow}
Chao, C.-H.; Sun, W.-F.; Hsu, Y.-C.; Kira, Z.; and Lee, C.-Y. 2023.
\newblock Training Energy-Based Normalizing Flow with Score-Matching
  Objectives.
\newblock arXiv:2305.15267.

\bibitem[{Chen et~al.(2018)Chen, Rubanova, Bettencourt, and
  Duvenaud}]{chen2018neuralode}
Chen, R. T.~Q.; Rubanova, Y.; Bettencourt, J.; and Duvenaud, D. 2018.
\newblock Neural Ordinary Differential Equations.
\newblock In \emph{Adv. Neural Inf. Proc. Sys.}, volume~31.

\bibitem[{Coeurdoux, Dobigeon, and Chainais(2022)}]{coeurdoux2022sliced}
Coeurdoux, F.; Dobigeon, N.; and Chainais, P. 2022.
\newblock Sliced-Wasserstein normalizing flows: beyond maximum likelihood
  training.
\newblock arXiv:2207.05468.

\bibitem[{Cramér and Wold(1936)}]{cramer1936some}
Cramér, H.; and Wold, H. 1936.
\newblock Some Theorems on Distribution Functions.
\newblock \emph{J. London Math. Soc.}, s1-11(4): 290--294.

\bibitem[{Cuturi(2013)}]{cuturi2013sinkhorn}
Cuturi, M. 2013.
\newblock Sinkhorn Distances: Lightspeed Computation of Optimal Transport.
\newblock In \emph{Adv. Neural Inf. Proc. Systems}, volume~26, 2292--2300.

\bibitem[{Deng et~al.(2026)Deng, Li, Li, Du, and He}]{deng2026drifting}
Deng, M.; Li, H.; Li, T.; Du, Y.; and He, K. 2026.
\newblock Generative Modeling via Drifting.
\newblock arXiv:2602.04770.

\bibitem[{Dinh, Sohl-Dickstein, and Bengio(2017)}]{dinh2016realnvp}
Dinh, L.; Sohl-Dickstein, J.; and Bengio, S. 2017.
\newblock Density estimation using Real NVP.
\newblock arXiv:1605.08803.

\bibitem[{Epps and Pulley(1983)}]{epps1983test}
Epps, T.~W.; and Pulley, L.~B. 1983.
\newblock A test of normality based on the empirical characteristic function.
\newblock \emph{Biometrika}, 70(3): 723--726.

\bibitem[{Genevay, Peyr{\'e}, and Cuturi(2018)}]{genevay2018learning}
Genevay, A.; Peyr{\'e}, G.; and Cuturi, M. 2018.
\newblock Learning Generative Models with Sinkhorn Divergences.
\newblock In \emph{Proc. 21st Int. Conf. Art. Intel. Stats.}, volume~84 of
  \emph{Proc. Machine Learning Research}, 1608--1617. PMLR.

\bibitem[{Golling et~al.(2023)Golling, Klein, Mastandrea, Nachman, and
  Raine}]{golling2023morphing}
Golling, T.; Klein, S.; Mastandrea, R.; Nachman, B.; and Raine, J.~A. 2023.
\newblock Morphing one dataset into another with max. likelihood estimation.
\newblock \emph{Phys. Rev. D}, 108(9): 096018.

\bibitem[{Grathwohl et~al.(2019)Grathwohl, Chen, Bettencourt, Sutskever, and
  Duvenaud}]{grathwohl2019ffjord}
Grathwohl, W.; Chen, R. T.~Q.; Bettencourt, J.; Sutskever, I.; and Duvenaud, D.
  2019.
\newblock {FFJORD}: Free-form Continuous Dynamics for Scalable Reversible
  Generative Models.
\newblock In \emph{Int. Conf. Learning Representations}.

\bibitem[{Gretton et~al.(2012)Gretton, Borgwardt, Rasch, Sch{\"o}lkopf, and
  Smola}]{gretton2012kernel}
Gretton, A.; Borgwardt, K.~M.; Rasch, M.~J.; Sch{\"o}lkopf, B.; and Smola, A.
  2012.
\newblock A Kernel Two-Sample Test.
\newblock \emph{J. Machine Learning Research}, 13(25): 723--773.

\bibitem[{Gretton et~al.(2026)Gretton, Wenliang, Galashov, Thornton, Bortoli,
  and Doucet}]{gretton2026drifting}
Gretton, A.; Wenliang, L.~K.; Galashov, A.; Thornton, J.; Bortoli, V.~D.; and
  Doucet, A. 2026.
\newblock On the Wasserstein Gradient Flow Interpretation of Drifting Models.
\newblock arXiv:2605.05118.

\bibitem[{Grover et~al.(2020)Grover, Chute, Shu, Cao, and
  Ermon}]{grover2020alignflow}
Grover, A.; Chute, C.; Shu, R.; Cao, Z.; and Ermon, S. 2020.
\newblock Alignflow: Cycle consistent learning from multiple domains via
  normalizing flows.
\newblock In \emph{Proc. AAAI Conf. Artificial Intelligence}, volume~34,
  4028--4035.

\bibitem[{Hull(1994)}]{hull1994database}
Hull, J.~J. 1994.
\newblock A Database for Handwritten Text Recognition Research.
\newblock \emph{IEEE Trans. Pattern Analysis and Machine Intelligence}, 16(5):
  550--554.

\bibitem[{Kerrigan, Migliorini, and Smyth(2024)}]{kerrigan2024functional}
Kerrigan, G.; Migliorini, G.; and Smyth, P. 2024.
\newblock Functional Flow Matching.
\newblock In \emph{Proc. 27th Int. Conf. Artificial Intelligence and
  Statistics}, volume 238 of \emph{Proceedings of Machine Learning Research},
  3934--3942. PMLR.

\bibitem[{Kingma and Dhariwal(2018)}]{kingma2018glow}
Kingma, D.~P.; and Dhariwal, P. 2018.
\newblock Glow: Generative Flow with Invertible 1x1 Convolutions.
\newblock arXiv:1807.03039.

\bibitem[{Knop et~al.(2020)Knop, Tabor, Spurek, Podolak, Mazur, and
  Jastrz{\k{e}}bski}]{knop2020cramerwold}
Knop, S.; Tabor, J.; Spurek, P.; Podolak, I.; Mazur, M.; and Jastrz{\k{e}}bski,
  S. 2020.
\newblock Cramer-Wold Auto-Encoder.
\newblock \emph{J. Machine Learning Research}, 21(164): 1--28.

\bibitem[{LeCun et~al.(1998)LeCun, Bottou, Bengio, and
  Haffner}]{lecun1998gradient}
LeCun, Y.; Bottou, L.; Bengio, Y.; and Haffner, P. 1998.
\newblock Gradient-Based Learning Applied to Document Recognition.
\newblock \emph{Proceedings of the IEEE}, 86(11): 2278--2324.

\bibitem[{Lipman et~al.(2023)Lipman, Chen, Ben-Hamu, Nickel, and
  Le}]{lipman2023flowmatching}
Lipman, Y.; Chen, R. T.~Q.; Ben-Hamu, H.; Nickel, M.; and Le, M. 2023.
\newblock Flow Matching for Generative Modeling.
\newblock In \emph{Int. Conf. Learning Representations}.

\bibitem[{Menze et~al.(2015)Menze, Jakab, Bauer, Kalpathy-Cramer, Farahani,
  Kirby, Burren, Porz, Slotboom, Wiest et~al.}]{menze2015multimodal}
Menze, B.~H.; Jakab, A.; Bauer, S.; Kalpathy-Cramer, J.; Farahani, K.; Kirby,
  J.; Burren, Y.; Porz, N.; Slotboom, J.; Wiest, R.; et~al. 2015.
\newblock The Multimodal Brain Tumor Image Segmentation Benchmark (BRATS).
\newblock \emph{IEEE Trans. Medical Imaging}, 34(10): 1993--2024.

\bibitem[{Papamakarios et~al.(2021)Papamakarios, Nalisnick, Rezende, Mohamed,
  and Lakshminarayanan}]{papamakarios2021normalizingflows}
Papamakarios, G.; Nalisnick, E.; Rezende, D.~J.; Mohamed, S.; and
  Lakshminarayanan, B. 2021.
\newblock Normalizing flows for probabilistic modeling and inference.
\newblock \emph{J. Machine Learning Research}, 22(57): 1--64.

\bibitem[{Pooladian et~al.(2023)Pooladian, Ben-Hamu, Domingo-Enrich, Amos,
  Lipman, and Chen}]{pooladian2023multisample}
Pooladian, A.-A.; Ben-Hamu, H.; Domingo-Enrich, C.; Amos, B.; Lipman, Y.; and
  Chen, R.~T. 2023.
\newblock Multisample flow matching: Straightening flows with minibatch
  couplings.
\newblock \emph{arXiv preprint arXiv:2304.14772}.

\bibitem[{Rabin et~al.(2011)Rabin, Peyr{\'e}, Delon, and
  Bernot}]{rabin2011wasserstein}
Rabin, J.; Peyr{\'e}, G.; Delon, J.; and Bernot, M. 2011.
\newblock Wasserstein Barycenter and Its Application to Texture Mixing.
\newblock In \emph{Scale Space and Variational Methods in Computer Vision},
  volume 6667, 435--446. Springer.

\bibitem[{Sabour, Fidler, and Kreis(2025)}]{sabour2025alignyourflow}
Sabour, A.; Fidler, S.; and Kreis, K. 2025.
\newblock Align Your Flow: Scaling Continuous-Time Flow Map Distillation.
\newblock arXiv:2506.14603.

\bibitem[{Wahl et~al.(2025)Wahl, Rousselot, Draxler, and
  Koethe}]{wahl2025trade}
Wahl, S.; Rousselot, A.; Draxler, F.; and Koethe, U. 2025.
\newblock {TRADE}: Transfer of Distributions between External Conditions with
  Normalizing Flows.
\newblock In \emph{Proc. 28th Int. Conf. Artificial Intelligence and
  Statistics}, volume 258 of \emph{Proceedings of Machine Learning Research},
  3196--3204. PMLR.

\bibitem[{Xie et~al.(2024)Xie, Zhu, Yu, Yang, Cheng, Zhang, Zhang, and
  Zhang}]{xie2024reflected}
Xie, T.; Zhu, Y.; Yu, L.; Yang, T.; Cheng, Z.; Zhang, S.; Zhang, X.; and Zhang,
  C. 2024.
\newblock Reflected Flow Matching.
\newblock In \emph{Proc. Int. Conf. Mac. Learn.}, volume 235, 54614--54634.

\bibitem[{Xiong, Ding, and Zhang(2025)}]{xiong2023symotflow}
Xiong, Z.; Ding, Q.; and Zhang, X. 2025.
\newblock Arbitrary Distributions Mapping via SyMOT-Flow: A Flow-Based Approach
  Integrating Maximum Mean Discrepancy and Optimal Transport.
\newblock \emph{SIAM J. Imaging Sciences}, 18(1): 570--596.

\bibitem[{Zimmermann et~al.(2025)Zimmermann, Wiltzer, Szeto, Alvarez-Melis, and
  Mackey}]{zimmermann2025kerjepa}
Zimmermann, E.; Wiltzer, H.; Szeto, J.; Alvarez-Melis, D.; and Mackey, L. 2025.
\newblock {KerJEPA}: Kernel Discrepancies for Euclidean Self-Supervised
  Learning.
\newblock arXiv:2512.19605.

\end{thebibliography}

\end{document}